\documentclass[conference]{IEEEtran}
\IEEEoverridecommandlockouts
\usepackage{cite}
\usepackage{amsmath,amssymb,amsfonts}
\usepackage{algorithmic}
\usepackage{graphicx}
\usepackage{textcomp}
\usepackage[ruled,vlined]{algorithm2e}
\usepackage{tikz}
\usepackage{enumitem} 
\usepackage{pgfplots}
\usepackage{pgfplotstable}
\usepackage{url}
\usepackage{amsmath} % 
\pgfplotsset{compat=newest}
\usepackage{booktabs}   % for \toprule, \midrule, \bottomrule
\usepackage{multirow}   % for \multirow
\usepackage{xcolor}
\def\BibTeX{{\rm B\kern-.05em{\sc i\kern-.025em b}\kern-.08em
    T\kern-.1667em\lower.7ex\hbox{E}\kern-.125emX}}
\usepackage[capitalize]{cleveref}

\begin{document}

\title{Joint Superpixel and Self-Representation Learning for Scalable Hyperspectral Image Clustering\\

\thanks{This work was supported by the Research Foundation – Flanders (G094122N, SPYDER), the Flanders AI Research Programme (174B09119), the Bijzonder Onderzoeksfonds (BOF.24Y.2021.0049.01), the China Scholarship Council (202106150007), the National Natural Science Foundation of China (42301425), and the China Postdoctoral Science Foundation (2023M743299).
}
}

\author{
    \IEEEauthorblockN{
        Xianlu Li\IEEEauthorrefmark{1}, 
        Nicolas Nadisic\IEEEauthorrefmark{1}\IEEEauthorrefmark{3}, 
        Shaoguang Huang\IEEEauthorrefmark{2} and
        Aleksandra  Pi\v{z}urica\IEEEauthorrefmark{1}
    }
    \IEEEauthorblockA{
        \IEEEauthorrefmark{1}Department of Telecommunications and Information Processing, Ghent University, Belgium.\\
    }
    \IEEEauthorblockA{
        \IEEEauthorrefmark{3}Royal Institute for Cultural Heritage (KIK-IRPA), Brussels, Belgium.\\
    }
    \IEEEauthorblockA{
        \IEEEauthorrefmark{2}School of Computer Science, China University of Geosciences, Wuhan, China.\\
    }
}
\maketitle

\begin{abstract}
Subspace clustering is a powerful unsupervised approach for hyperspectral image (HSI) analysis, but its high computational and memory costs limit scalability. Superpixel segmentation can improve efficiency by reducing the number of data points to process. However, existing superpixel-based methods usually perform segmentation independently of the clustering task, often producing partitions that do not align with the subsequent clustering objective. To address this, we propose a unified end-to-end framework that jointly optimizes superpixel segmentation and subspace clustering. Its core is a feedback mechanism: a self-representation network based on unfolded Alternating Direction Method of Multipliers (ADMM) provides a model-driven signal to guide a differentiable superpixel module. This joint optimization yields “clustering-aware” partitions that preserve spectral–spatial structure. Furthermore, our superpixel network learns a unique compactness parameter for each superpixel, enabling more flexible and adaptive segmentation. Extensive experiments on benchmark HSI datasets demonstrate that our method consistently achieves superior accuracy compared with state-of-the-art clustering approaches.

\end{abstract}


\section{Introduction}
Hyperspectral images (HSI) capture rich spectral information across hundreds of contiguous bands, enabling fine-grained material discrimination. As a result, HSI has been widely applied in domains such as environmental monitoring and precision agriculture~\cite{khan2018modern}. HSI clustering, which groups pixels into distinct categories in an unsupervised manner, is critical in HSI analysis due to the frequent lack of labeled data~\cite{huang2023model}. A variety of clustering methods have been explored, including centroid-based~\cite{Kmeans,FCM}, graph-based~\cite{spectral_clustering}, hierarchical~\cite{finch}, and subspace-based approaches~\cite{SSC,9081919}.

Among these, subspace-based clustering has attracted increasing attention for its ability to model data as unions of low-dimensional subspaces. Representative methods include sparse subspace clustering (SSC)~\cite{SSC, 8451277, zhang2016spectral}, low-rank representation (LRR)~\cite{9081919}, and deep subspace clustering~\cite{DSCNet,cai2021hypergraph, li2021self}. However, the computational and memory requirements of these methods scale quadratically with the number of pixels, severely limiting their scalability.

To reduce this complexity, several superpixel-based subspace clustering methods have been proposed to enhance spatial consistency and efficiency~\cite{cai2022superpixel,9641802}. Yet, most existing approaches generate superpixels as a preprocessing step independent of the clustering objective~\cite{cai2022superpixel}, often leading to partitions misaligned with the downstream clustering goal and thus limiting overall performance.

While prior works like Superpixel Sampling Networks (SSN)~\cite{jampani18ssn} introduced differentiable superpixel frameworks, their application has been mainly in supervised tasks where gradients are derived from ground-truth labels. Extending this idea to unsupervised clustering is challenging because no ground-truth labels are available to provide gradient signals. To overcome this, we employ the self-representation objective as a model-driven signal to guide superpixel formation. This yields clustering-aware partitions that align with the intrinsic data structure and provide a cleaner, more structured input to the clustering module. As a result, our method achieves both high efficiency and strong clustering accuracy. Our main contributions are as follows:
\begin{itemize}
    \item We propose a unified and efficient framework that jointly optimizes superpixel generation and self-representation learning, enabling clustering-oriented superpixels that align with the downstream clustering objective.
    \item We introduce a differentiable superpixel generation network with learnable compactness parameters, where each superpixel adaptively balances spectral and spatial characteristics for more flexible and structure-preserving segmentation.
    \item We design structural constraints across multiple granularities, including pixel--pixel local consistency, pixel--superpixel compactness, and superpixel--superpixel representation, which together preserve spectral--spatial structure and improve clustering performance.
\end{itemize}

\section{METHODOLOGY}
\label{METHODOLOGY}

\begin{figure}[ht]
    \centering
    \includegraphics[width=0.8\linewidth]{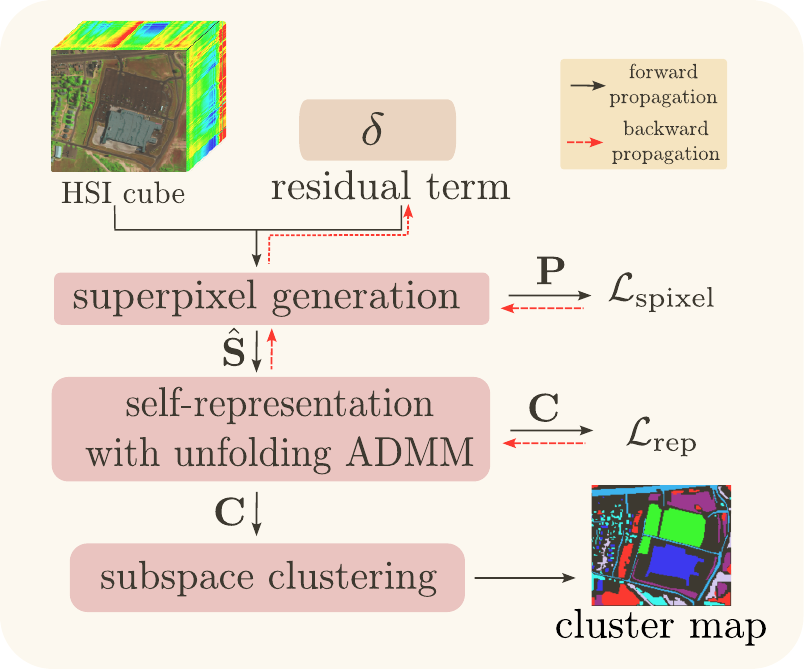}
    \caption{Structure of the proposed method.}
    \label{fig:framework}
\end{figure}

In this section, we present our framework that jointly optimizes superpixel segmentation and self-representation. It consists of two differentiable modules: superpixel generation and self-representation, coupled in a unified training loop. The structural feedback from the clustering module guides superpixel formation, making them not only visually coherent but also aligned with the intrinsic cluster structure. An overview is shown in Fig.~\ref{fig:framework}.

\subsection{Problem Formulation}

Let $\mathbf{X} \in \mathbb{R}^{N \times D}$ denote a hyperspectral image (HSI), where $N$ is the number of pixels and $D$ the number of spectral bands. The objective of HSI clustering is to assign each pixel to one of $\mathcal{C}$ clusters in an unsupervised manner. Subspace-based clustering methods, which rely on the self-representation principle that each data point can be represented as a linear combination of others within the same subspace~\cite{SSC}, have demonstrated strong performance but suffer from quadratic complexity with respect to the number of pixels. Superpixels reduce clustering units, improving efficiency while preserving spectral–spatial structure. Yet, most existing approaches generate superpixels as a preprocessing step independent of the clustering objective, often leading to partitions misaligned with the downstream clustering goal and thus limiting overall performance.

To obtain clustering-aware superpixels and thereby improve clustering accuracy and efficiency, we define the following joint end-to-end optimization, where the self-representation structure provides feedback to refine superpixels:
\begin{equation}
\label{target_function}
\min_{\mathbf{P},\,\mathbf{S},\,\mathbf{C}}\;
\varphi(\mathbf{X}',\mathbf{P},\mathbf{S}) \;+\; \eta(\hat{\mathbf{S}},\mathbf{C}), \quad \text{s.t. } \mathrm{diag}(\mathbf{C})=\mathbf{0},
\end{equation}
where $\varphi$ and $\eta$ denote the superpixel and self-representation objectives, detailed in the following subsections. 

Here, $\mathbf{X}' = \mathbf{X} + \boldsymbol{\delta}$ denotes the adapted feature matrix, where $\boldsymbol{\delta} \in \mathbb{R}^{N \times D}$ is a learnable residual term. This residual allows the network to slightly adjust the original features during training, so that the resulting representation becomes more discriminative for clustering and superpixel segmentation. The matrix $\mathbf{P} \in \mathbb{R}^{N \times M}$ denotes the soft assignment of $N$ pixels to $M$ superpixels, 
whose spectral centroids $\mathbf{S} \in \mathbb{R}^{M \times D}$ are computed as the $\mathbf{P}$-weighted means of $\mathbf{X}'$. 
We then obtain $\hat{\mathbf{S}}$ by applying column-wise $\ell_2$ normalization to $\mathbf{S}^\top$, 
and use it to estimate the coefficient matrix $\mathbf{C} \in \mathbb{R}^{M \times M}$, which encodes sparse self-representation relations such that each superpixel is linearly reconstructed by only a few others.

To optimize Eq.~\eqref{target_function}, we first introduce a superpixel generation network, where pixel-level assignment is made differentiable to jointly optimize $\varphi$ with the clustering objective, and then employ an unfolded ADMM-based self-representation network to solve $\eta$, providing structural feedback to the superpixel generation.

\subsection{Superpixel Generation Network}
A limitation of conventional superpixel methods is that they rely on a single, fixed compactness parameter, which cannot adapt to the diverse spectral–spatial trade-offs required across different regions of an image or to the needs of the downstream clustering task. To overcome this, we introduce a learnable weight vector $\mathbf{W}$, where each element $w_j$ controls the compactness of the $j$-th superpixel and is optimized jointly with the clustering objective. This design enables superpixels to adapt their balance between spectral compactness and spatial coherence in a task-driven manner, resulting in more flexible and structure-preserving partitions. The distance $d_{ij}$ between the $i$-th pixel and the $j$-th superpixel is defined as:
\begin{equation}
\label{eq:distance}
\begin{split}
d_{ij} =\; & w_j \cdot \left\| \mathbf{X'}_i - \mathbf{S}_j \right\|^2 \\
& + (1 - w_j) \cdot \left\| r(\mathbf{X'}_i) - r(\mathbf{S}_j) \right\|^2 ,
\end{split}
\end{equation}
where $w_j$ is the learnable weight to balance feature similarity and spatial compactness in the $j_{th}$ superpixel. The $r(\cdot)$ denotes the spatial coordinate operator for both pixels and superpixel centroids: $r(\mathbf{X'}_i)$ gives the pixel location. At the beginning of the process, the superpixel centers $\mathbf{S}_j$ and their spatial coordinates $r(\mathbf{S}_j)$ are initialized by a simple grid partitioning (or random assignment). The soft-assignment probabilities are obtained by a softmax:
\begin{equation}
\label{eq:softmax}
p_{ij} = \frac{\exp(-d_{ij} / \tau)}{\sum_{n\in \mathcal{P}_i} \exp(-d_{in} / \tau)},
\end{equation}
where $\tau$ controls the sharpness of the distribution, and $\mathcal{P}_i$ denotes the $G$ nearest superpixels to pixel $i$ 
(fixed to $G{=}9$ by default in~\cite{achanta2012slic,jampani18ssn}). This softmax-based assignment is fully differentiable, enabling gradients to flow through the superpixel generation process and thus supporting end-to-end training. Based on these assignments, the superpixel centers are updated 
by a weighted average of the assigned pixels:
\begin{equation}
\label{eq:joint_centers_single}
\begin{bmatrix} \mathbf{S}_j \\[2pt] r(\mathbf{S}_j) \end{bmatrix}
= \frac{1}{\sum_{i=1}^N p_{ij}}
\sum_{i=1}^N p_{ij}
\begin{bmatrix} \mathbf{X}'_i \\[2pt] r(\mathbf{X}'_i) \end{bmatrix},
\quad j=1,\dots,M.
\end{equation}
The assignments and superpixel centers are refined iteratively, as summarized in Algorithm~\ref{alg:ssn_iter}.
\begin{algorithm}[ht]
\caption{Iterative Superpixel Assignment}
\label{alg:ssn_iter}
\KwIn{
    Spectral feature matrix $\mathbf{X}' \in \mathbb{R}^{N \times D}$, \\
    Number of superpixels $M$, number of iterations $T$, \\
    compactness $\mathbf{W}$, temperature $\tau$
}
\KwOut{Soft assignment matrix $\mathbf{P}$, hard labels $\mathbf{L}$}

\textbf{Step 1: Initialization}\;
Initialize $[\mathbf{S}, r(\mathbf{S})]$ by grid partitioning\;

\textbf{Step 2: Iterative refinement}\;
\For{$t = 1$ \KwTo $T$}{
    Compute distance matrix $d_{ij}$ using \eqref{eq:distance}\; 
    Obtain the soft-assignment matrix using \eqref{eq:softmax}\;   
    Update superpixel centers $[\mathbf{S}^t, r(\mathbf{S}^t)]$ using \eqref{eq:joint_centers_single};
}

\textbf{Step 3: Hard assignment}\;
Derive hard labels $\mathbf{L}$ from soft assignment matrix $\mathbf{P}$\;

\Return{$\mathbf{P}$, $\mathbf{L}$}
\end{algorithm}

To obtain high-quality superpixels, we define $\varphi(\mathbf{X}',\mathbf{P},\mathbf{S})$ as the superpixel loss $\mathcal{L}_{\text{spixel}}$:
\begin{equation}
\mathcal{L}_{\text{spixel}}
=
\underbrace{\frac{1}{N} \sum_{i=1}^{N} \left\| \mathbf{X}'_i - \mathbf{F}_i \right\|^2}_{\text{spectral compactness}}
+
\underbrace{\sum_{i=1}^{N} \sum_{n \in \mathcal{N}(i)} \big( 1 - \mathrm{sim}(\mathbf{P}_i, \mathbf{P}_n) \big)}_{\text{local consistency}} .
\end{equation}
Here, $\mathbf{F}\in\mathbb{R}^{N\times D}$ is obtained by assigning each pixel 
to its superpixel centroid using $\mathbf{S}$ and the label map $\mathbf{L}$, 
so that $\mathbf{F}_i$ denotes the centroid of the superpixel containing pixel $i$. 
The first term enforces intra-superpixel compactness by pulling each pixel feature 
$\mathbf{X}'_i$ toward $\mathbf{F}_i$, enhancing homogeneity within superpixels. 
The second term encourages neighboring pixels to share the same superpixel label, 
reducing noise and yielding spatially coherent partitions. Together, these terms 
preserve both pixel–pixel and pixel–superpixel structure.

\subsection{Self-representation Network (Unfolded ADMM)}
After obtaining superpixels, their pairwise similarities are obtained via sparse self-representation. Following the formulation of Sparse Subspace Clustering (SSC)~\cite{SSC}, we define $\eta(\hat{\mathbf{S}},\mathbf{C})$ as:
\begin{equation}
\label{eq:self_representation}
\eta(\hat{\mathbf{S}},\mathbf{C})
= \big\|\hat{\mathbf{S}}-\hat{\mathbf{S}}\mathbf{C}\big\|_{F}^{2}
+ \lambda_{\text{sr}} \|\mathbf{C}\|_{1},
\end{equation}
where the first term enforces accurate reconstruction among normalized superpixel features $\hat{\mathbf{S}}$, the second term imposes sparsity on the coefficient matrix $\mathbf{C}$. The diagonal constraint $\mathrm{diag}(\mathbf{C})=0$, already enforced in the overall optimization problem~\eqref{target_function}, prevents trivial self-representation. In classical SSC, $\lambda_{\text{sr}}$ is fixed. 
In unfolded formulations, however, $\lambda_{\text{sr}}$ is treated as a learnable parameter, so that the sparsity level can adapt during training.

To facilitate optimization, we introduce an auxiliary variable \( \mathbf{Z} \), which leads to the equivalent constrained formulation:
\begin{equation}
  \label{eq:split-problem}
  \min_{\mathbf C,\mathbf Z}\;
  \lVert \hat{\mathbf{S}}  - \hat{\mathbf{S}}\mathbf C\rVert_F^{2}
  +\lambda_{\text{sr}}\lVert\mathbf Z\rVert_{1},
  \quad
  \text{s.t. } \mathbf C=\mathbf Z,\;
                \operatorname{diag}(\mathbf Z)=\mathbf 0.
\end{equation}
To solve this constrained problem, we form the following augmented Lagrangian:
\begin{equation}
  \label{eq:lagrangian}
  \mathcal L(\mathbf C,\mathbf Z,\mathbf \mu)=
  \lVert \hat{\mathbf{S}} - \hat{\mathbf{S}}\mathbf C\rVert_F^{2}
  +\lambda_{\text{sr}}\lVert\mathbf Z\rVert_{1}
  +\langle\mathbf \mu,\mathbf C-\mathbf Z\rangle
  +\frac{\rho}{2}\lVert\mathbf C-\mathbf Z\rVert_{F}^{2},
\end{equation}
where $\mathbf \mu$ is the dual variable and $\rho>0$ is the penalty parameter.  
The optimization proceeds with iterative updates:\\
\textbf{C‑update:}

\begin{equation}\label{eq:C-update}
\mathbf{C}^{k+1} = (2\hat{\mathbf{S}}^T\hat{\mathbf{S}}+\rho I)^{-1} \Big( 2\hat{\mathbf{S}}^T\hat{\mathbf{S}} - (\mu^{k} - \rho \mathbf{Z}^{k}) \Big).
\end{equation}

\textbf{Z‑update:}
  \begin{equation}
         \mathbf{Z}^{(k+1)} = \operatorname{ReLU}(|\mathbf{C}^{k+1} + \frac{\mu^{k}}{\rho}| - \frac{\lambda_{\text{sr}}^{k}}{\rho})\cdot \text{sgn}(\mathbf{C}^{k+1} + \frac{\mu^{k}}{\rho}),
    \end{equation}
    \begin{equation}
        \mathbf{Z}^{(k+1)} = \mathbf{Z}^{(k+1)} - \operatorname{diag}(\mathbf{Z}^{(k+1)}),
    \end{equation}
where $\operatorname{ReLU}(\cdot) = \max(0, \cdot)$ is the standard rectified linear unit for the element-wise soft-thresholding.\\
\textbf{$\mu$‑update:}
\begin{equation}
\label{eq:u-update}
\mathbf \mu^{(k+1)} = \mathbf \mu^{(k)}+\rho\bigl(\mathbf C^{(k+1)}-\mathbf Z^{(k+1)}\bigr).
\end{equation}
Through the differentiable ReLU-based thresholding and the introduction of learnable $\lambda_{sr}$, the ADMM updates are unfolded into a finite-depth neural network, where each iteration corresponds to one layer. This formulation embeds the optimization process into the network architecture and enables end-to-end training. According to the self-representation objective defined in Eq.~\eqref{eq:self_representation}, we impose the following losses.

\textbf{Reconstruction loss.}
\begin{equation}
\mathcal{L}_{\text{recon}} = \left\lVert \hat{\mathbf{S}}\mathbf{C} - \hat{\mathbf{S}} \right\rVert_F^2,
\end{equation}
which enforces accurate reconstruction of superpixels using the self-representation matrix. 

\textbf{Sparsity regularization.}
We further constrain the coefficient matrix $\mathbf{C}$ with two complementary terms:
\begin{equation}
\mathcal{L}_{\ell_1} = \|\mathbf{C}\|_1, \quad
\mathcal{L}_{\text{entropy}} = - \frac{1}{M} \sum_{j=1}^{M} \sum_{i=1}^{M} \tilde{c}_{ij}\log(\tilde{c}_{ij}+\varepsilon),
\end{equation}
where $\tilde{c}_{ij}$ is the normalized absolute value of $c_{ij}$, and $\epsilon$ is a stability constant, empirically set to $10^{-8}$. The $\ell_1$ term suppresses weak connections, and the entropy term promotes peaked distributions and improves clustering robustness. The final objective of the self-representation module for uncovering superpixel level structure is formulated as a weighted sum of $\mathcal{L}_{\text{recon}}$, $\mathcal{L}_{\ell_1}$, $\mathcal{L}_{\text{entropy}}$. For simplicity, we set:
\begin{equation}
\mathcal{L}_{\text{rep}} = 2\cdot\mathcal{L}_{\text{recon}} + \mathcal{L}_{\ell_1} + \mathcal{L}_{\text{entropy}},
\end{equation}
and refer to it as the self-representation loss. 
\subsection{Complete Loss Function}
The overall training objective integrates the self-representation loss, superpixel regularization, and a noise term, providing multi-granularity constraints, defined as:
\begin{equation}
    \mathcal{L}_{\text{all}} = \alpha \cdot \mathcal{L}_{\text{rep}} + \mathcal{L}_{\text{spixel}} + \mathcal{L}_{\text{noise}} ,
\end{equation}
where \( \mathcal{L}_{\text{noise}} \) regularizes the learnable residual term \( \boldsymbol{\delta} \) added to the input features. We define it as:
\begin{equation}
    \mathcal{L}_{\text{noise}} = \frac{\lambda}{ND} \left\| \boldsymbol{\delta} \right\|_F^2 ,
\end{equation}
where $ND$ is the size of $\boldsymbol{\delta}$ and $\lambda$ is a parameter. In all experiments, we fix $\lambda=50$. The only tunable parameter is $\alpha$, whose effect will be examined in the next section, as its value depends on the dataset.

% In all our experiments, we fixed $\lambda=50$ and only the parameter $\alpha$ remains tunable and will be analyzed in the parameter study in the next section, as its value depends on the dataset.

\section{Experiments}
To demonstrate the effectiveness of the proposed method, we evaluate its performance through experiments on widely used benchmark hyperspectral datasets. All implementation details are provided in our source code\footnote{\url{https://github.com/lxlscut/superpixel_ssc}}.

\subsection{Experiment Setup}
In this section, we evaluate the proposed method on three benchmark datasets described in \cref{tab:datasets}. 

\begin{table}[ht]
\centering
\caption{Summary of the HSI datasets used in the experiments.}
\label{tab:datasets}
\resizebox{0.9\linewidth}{!}{
\begin{tabular}{lcccc}
\toprule
\textbf{Dataset} & \textbf{Bands} & \textbf{Size (pixels)} & \textbf{Classes} & \textbf{Labeled Samples} \\
\midrule
Trento  & 63  & $600 \times 166$ & 6  & 30,214 \\
Salinas & 204 & $512 \times 217$ & 16 & 54,129 \\
Urban   & 162 & $150 \times 160$  & 7  & 12,048 \\
\bottomrule
\end{tabular}}
\end{table}

The number of superpixels is set as the maximum of two values: (1) the edge-based estimation in~\cite{cai2022superpixel}, which reflects scene complexity, and (2) a lower bound $50\mathcal{C}/\mathcal{R}$, where $\mathcal{C}$ is the number of classes and $\mathcal{R}$ is the area ratio of the cluster region to the entire image. The factor 50 is an empirical choice to ensure sufficient samples per class~\cite{SSC}. This dual criterion makes the number of superpixels both adaptive to spatial complexity and robust across different class distributions.

\subsection{Results}
\subsubsection{Comparison with other methods}
In our experiments, we compare the proposed method with several widely adopted clustering approaches, including the centroid-based methods FCM~\cite{FCM}, DEKM~\cite{DEKM}, the graph-based methods SpectralNet~\cite{SpectralNet}, the hierarchical method FINCH~\cite{finch}, the deep clustering method IDEC~\cite{guo2017improved}, and the basis-representation-based subspace clustering methods MADL~\cite{li2023model}. In addition, we include a comparison with the superpixel-based subspace clustering method NCSC~\cite{cai2022superpixel}. Specifically, the source codes for IDEC~\cite{guo2017improved} and DEKM~\cite{DEKM} are obtained from the ClustPy library~\cite{leiber2023benchmarking}, while the implementations of the other recent methods are taken from their official repositories. 
The results of the experiments are shown in \cref{tab:clustering_results}.

\begin{table}[ht]
\centering
\caption{Clustering performance comparison results.}
\label{tab:clustering_results}
\resizebox{0.48\textwidth}{!}{%
\begin{tabular}{lccc|ccc|ccc}
\toprule
\multirow{2}{*}{\textbf{Method}} & \multicolumn{3}{c|}{\textbf{Salinas}} & \multicolumn{3}{c|}{\textbf{Trento}} & \multicolumn{3}{c}{\textbf{Urban}} \\
& OA (\%) & NMI & $\kappa$ & OA (\%) & NMI & $\kappa$ & OA (\%) & NMI & $\kappa$ \\
\midrule
K-means     & 71.43 & 0.7829 & 0.6840 & 83.54 & 0.7439 & 0.7829 & 72.59 & 0.6932 & 0.6733 \\
FCM         & 65.57 & 0.7597 & 0.6257 & 75.63 & 0.7191 & 0.6909 & \underline{78.06} & 0.7065 & 0.7397 \\
FINCH       & 65.63 & 0.7941 & 0.6099 & 81.37 & 0.7932 & 0.7457 & 72.23 & 0.7060 & 0.6728 \\
DEKM        & \underline{74.92} & \textbf{0.8635} & 0.7176 & 85.48 & 0.8047 & 0.8018 & 74.30 & \underline{0.7454} & 0.6933 \\
SpectralNet & 69.87 & 0.7904 & 0.6707 & 68.35 & 0.7764 & 0.5982 & 69.13 & 0.7183 & 0.6326 \\
IDEC        & 53.48 & 0.7188 & 0.4949 & 69.69 & 0.7155 & 0.6246 & 70.22 & 0.6804 & 0.6503 \\
MADL        & 51.95 & 0.6928 & 0.4560 & \underline{88.72} & \textbf{0.9292} & \underline{0.8490} & 72.49 & 0.7166 & 0.6740 \\
NCSC        & 55.29 & 0.6409 & 0.5066 & 66.17 & 0.5160 & 0.5548 & 53.61 & 0.4742 & 0.4481 \\
Ours & \textbf{80.96} & \underline{0.8354} & \textbf{0.7868} & \textbf{90.21} & \underline{0.8297} & \textbf{0.8694} & \textbf{84.52} & \textbf{0.7517} & \textbf{0.8173} \\
\bottomrule
\end{tabular}
}
\end{table}

The baseline methods show varying strengths: DEKM performs well on Salinas, while MADL achieves strong results on Trento but fails on Salinas. The superpixel-based NCSC lags behind across all datasets. In contrast, our method consistently attains the highest OA and $\kappa$, while also ranking among the top in NMI, demonstrating clear advantages in accuracy, robustness, and generalization.

\subsubsection{Ablation Study}
To assess the contribution of the self-representation module, we conduct an ablation study with four model variants: M1, direct clustering on initial superpixels without training; M2, training only the superpixel generation network with $\mathcal{L}_{\text{spixel}}$ and $\mathcal{L}_{\text{noise}}$; M3, training only the unfolded ADMM network with $\mathcal{L}_{\text{rep}}$; M4, training superpixel generation network and unfolded ADMM network separately; and Ours, the full model jointly optimized with all three losses. The corresponding results are presented in Table~\ref{tab:ablation}.
\begin{table}[ht]
\centering
\caption{Ablation study results.}
\label{tab:ablation}
\resizebox{\linewidth}{!}{
\begin{tabular}{lccc|ccc|ccc}
\toprule
\multirow{2}{*}{Model Variant} 
  & \multicolumn{3}{c|}{Salinas} 
  & \multicolumn{3}{c|}{Trento} 
  & \multicolumn{3}{c}{Urban} \\
  & OA (\%) & NMI & $\kappa$ 
  & OA (\%) & NMI & $\kappa$ 
  & OA (\%) & NMI & $\kappa$ \\
\midrule
M1             
  & 78.90 & 0.8220 & 0.7639
  & 85.09 & 0.7884 & 0.8011
  & 80.95 & 0.7239 & 0.7746 \\
M2             
  & 77.86 & 0.8251 & 0.7523
  & 87.92 & 0.8045 & 0.8384
  & 81.11 & 0.7243 & 0.7765 \\
M3             
  & 79.60 & 0.8332 & 0.7716
  & 85.41 & 0.8119 & 0.8054
  & 80.73 & 0.7234 & 0.7719 \\
M4  
& 77.51 & 0.8212 & 0.7483
& 87.94 & 0.7972 & 0.8387
& 80.82 & 0.7203 & 0.7729 \\
Full model (Ours)                                
  & \textbf{80.96} & \textbf{0.8354} & \textbf{0.7868}
  & \textbf{90.21} & \textbf{0.8297} & \textbf{0.8694}
  & \textbf{84.52} & \textbf{0.7517} & \textbf{0.8173} \\
\bottomrule
\end{tabular}}
\end{table}

% The ablation results show that neither the superpixel module (M2) nor the self-representation module (M3) alone achieves stable gains. Only the joint optimization in our full model consistently improves performance across datasets, indicating that the self-representation module not only enhances feature representation but also guides superpixel refinement.

The ablation results show that only the joint optimization in our full model get the best performance. This indicates that the joint optimization yields clustering-oriented superpixels, which in turn increase clustering performance across datasets.

\subsection{Parameter Study}

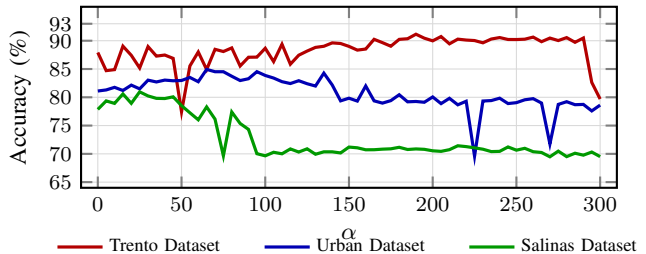
\begin{figure}[htbp]
    \centering
    \begin{tikzpicture}[trim axis left, trim axis right]
        \begin{axis}[
            width=0.48\textwidth,
            height=0.22\textwidth,
            xlabel={$\alpha$},
            ylabel={Accuracy (\%)},
            grid=major,
            grid style={gray!30},
            xtick={0,50,100,150,200,250,300},
            ytick={50,55,60,65,70,75,80,85,90,93},
            ymin=64, ymax=96,
            xmin=-10, xmax=310,
            legend style={
                at={(0.5,-0.22)}, 
                anchor=north, 
                legend columns=3,
                draw=none,
                fill=none,
                font=\scriptsize,
                /tikz/every even column/.append style={column sep=0.5cm}
            },
            axis line style={thick},
            tick style={thick},
            label style={font=\small},
            tick label style={font=\footnotesize},
            % 添加调试选项
            scaled y ticks=false,
            y tick label style={/pgf/number format/fixed},
        ]
        
        1. Trento Dataset
        \addplot[
            color=red!70!black,
            % mark=*,
            mark size=2.5pt,
            thick,
            line width=1.2pt,
            mark options={fill=red!70!black}
        ]
        table [x=weight,     y expr = {\thisrow{accuracy} * 100}, col sep=space] {weight_vs_accuracy_trento.dat};
        \addlegendentry{Trento Dataset}
        
        % 2. Urban Dataset  
        \addplot[
            color=blue!70!black,
            % mark=square*,
            mark size=2.5pt,
            thick,
            line width=1.2pt,
            mark options={fill=blue!70!black}
        ]
        table [x=weight, y expr = {\thisrow{accuracy} * 100}, col sep=space] {weight_vs_accuracy_urban.dat};
        \addlegendentry{Urban Dataset}
        
        % 3. Salinas Dataset
        \addplot[
            color=green!60!black,
            % mark=triangle*,
            mark size=3pt,
            thick,
            line width=1.2pt,
            mark options={fill=green!60!black}
        ]
        table [x=weight, y expr = {\thisrow{accuracy} * 100}, col sep=space] {weight_vs_accuracy_salinas.dat};
        \addlegendentry{Salinas Dataset}

    \end{axis}
    \end{tikzpicture}
    \caption{Impact of $\alpha$ on classification accuracy.}
    \label{papameter_visualize}
\end{figure}

We evaluate the impact of the self-representation loss weight $\alpha$ on clustering performance, as illustrated in Fig.~\ref{papameter_visualize}.

As shown in Fig.~\ref{papameter_visualize}, the proposed method demonstrates strong robustness to the self-representation loss weight $\alpha$ on Trento, where accuracy stabilizes above 89\% once $\alpha > 180$ and only declines at very large values. On Urban, accuracy improves gradually, peaks between 65 and 100, and then decreases. By contrast, Salinas is more sensitive, reaching optimal performance between 25 and 45. These results suggest that simpler datasets, which already yield strong baseline performance, tolerate larger $\alpha$ and show greater robustness, while more complex datasets require smaller $\alpha$ values and are more sensitive to parameter variation.

\section{Conclusion}
\label{sec:conclu}
% In this work, we introduced a unified framework that jointly optimizes superpixel generation and self-representation for hyperspectral image clustering. By coupling a differentiable superpixel module with an unfolded ADMM network, the model produces clustering-aware partitions that enhance representation quality. Experiments on multiple benchmarks confirm consistent improvements over existing methods, demonstrating the effectiveness and robustness of the proposed approach.
In this work, we introduced a unified framework that jointly optimizes superpixel generation and self-representation for hyperspectral image clustering. By coupling a differentiable superpixel module with an unfolded ADMM network through structural feedback, the model produces clustering-aware partitions that improve representation quality. The framework further integrates multi-granularity constraints for spectral--spatial structure preservation. Experiments on multiple benchmarks confirm consistent improvements over existing methods, demonstrating the effectiveness and robustness of the proposed approach.

\bibliographystyle{IEEEtran}
\bibliography{reference}

\begin{thebibliography}{10}
\providecommand{\url}[1]{#1}
\csname url@samestyle\endcsname
\providecommand{\newblock}{\relax}
\providecommand{\bibinfo}[2]{#2}
\providecommand{\BIBentrySTDinterwordspacing}{\spaceskip=0pt\relax}
\providecommand{\BIBentryALTinterwordstretchfactor}{4}
\providecommand{\BIBentryALTinterwordspacing}{\spaceskip=\fontdimen2\font plus
\BIBentryALTinterwordstretchfactor\fontdimen3\font minus \fontdimen4\font\relax}
\providecommand{\BIBforeignlanguage}[2]{{%
\expandafter\ifx\csname l@#1\endcsname\relax
\typeout{** WARNING: IEEEtran.bst: No hyphenation pattern has been}%
\typeout{** loaded for the language `#1'. Using the pattern for}%
\typeout{** the default language instead.}%
\else
\language=\csname l@#1\endcsname
\fi
#2}}
\providecommand{\BIBdecl}{\relax}
\BIBdecl

\bibitem{khan2018modern}
M.~J. Khan, H.~S. Khan, A.~Yousaf, K.~Khurshid, and A.~Abbas, ``Modern trends in hyperspectral image analysis: A review,'' \emph{Ieee Access}, vol.~6, pp. 14\,118--14\,129, 2018.

\bibitem{huang2023model}
S.~Huang, H.~Zhang, H.~Zeng, and A.~Pi{\v{z}}urica, ``From model-based optimization algorithms to deep learning models for clustering hyperspectral images,'' \emph{Remote Sensing}, vol.~15, no.~11, p. 2832, 2023.

\bibitem{Kmeans}
J.~A. Hartigan and M.~A. Wong, ``Algorithm as 136: A k-means clustering algorithm,'' \emph{Journal of the {R}oyal {S}tatistical {S}ociety. series {C} (applied statistics)}, vol.~28, no.~1, pp. 100--108, 1979.

\bibitem{FCM}
J.~C. Bezdek, \emph{Pattern recognition with fuzzy objective function algorithms}.\hskip 1em plus 0.5em minus 0.4em\relax {Springer Science \& Business Media}, 2013.

\bibitem{spectral_clustering}
U.~Von~Luxburg, ``A tutorial on spectral clustering,'' \emph{Statistics and computing}, vol.~17, no.~4, pp. 395--416, 2007.

\bibitem{finch}
S.~Sarfraz, V.~Sharma, and R.~Stiefelhagen, ``Efficient parameter-free clustering using first neighbor relations,'' in \emph{{Proceedings of the IEEE/CVF conference on computer vision and pattern recognition}}, 2019, pp. 8934--8943.

\bibitem{SSC}
E.~Elhamifar and R.~Vidal, ``Sparse subspace clustering: Algorithm, theory, and applications,'' \emph{{IEEE} {T}ransactions on {P}attern {A}nalysis and {M}achine {I}ntelligence}, vol.~35, no.~11, pp. 2765--2781, 2013.

\bibitem{9081919}
J.~Xu, J.~E. Fowler, and L.~Xiao, ``Hypergraph-regularized low-rank subspace clustering using superpixels for unsupervised spatial–spectral hyperspectral classification,'' \emph{IEEE Geoscience and Remote Sensing Letters}, vol.~18, no.~5, pp. 871--875, 2021.

\bibitem{8451277}
S.~Huang, H.~Zhang, and A.~Pižurica, ``Joint sparsity based sparse subspace clustering for hyperspectral images,'' in \emph{2018 25th IEEE International Conference on Image Processing (ICIP)}, 2018, pp. 3878--3882.

\bibitem{zhang2016spectral}
H.~Zhang, H.~Zhai, L.~Zhang, and P.~Li, ``Spectral--spatial sparse subspace clustering for hyperspectral remote sensing images,'' \emph{IEEE Transactions on Geoscience and Remote Sensing}, vol.~54, no.~6, pp. 3672--3684, 2016.

\bibitem{DSCNet}
P.~Ji, T.~Zhang, H.~Li, M.~Salzmann, and I.~Reid, ``Deep subspace clustering networks,'' \emph{{A}dvances in {N}eural {I}nformation {P}rocessing {S}ystems}, vol.~30, 2017.

\bibitem{cai2021hypergraph}
Y.~Cai, Z.~Zhang, Z.~Cai, X.~Liu, and X.~Jiang, ``Hypergraph-structured autoencoder for unsupervised and semisupervised classification of hyperspectral image,'' \emph{IEEE Geoscience and Remote Sensing Letters}, vol.~19, pp. 1--5, 2021.

\bibitem{li2021self}
K.~Li, Y.~Qin, Q.~Ling, Y.~Wang, Z.~Lin, and W.~An, ``Self-supervised deep subspace clustering for hyperspectral images with adaptive self-expressive coefficient matrix initialization,'' \emph{IEEE Journal of Selected Topics in Applied Earth Observations and Remote Sensing}, vol.~14, pp. 3215--3227, 2021.

\bibitem{cai2022superpixel}
Y.~Cai, Z.~Zhang, P.~Ghamisi, Y.~Ding, X.~Liu, Z.~Cai, and R.~Gloaguen, ``Superpixel contracted neighborhood contrastive subspace clustering network for hyperspectral images,'' \emph{IEEE Transactions on Geoscience and Remote Sensing}, vol.~60, pp. 1--13, 2022.

\bibitem{9641802}
H.~Zhao, F.~Zhou, L.~Bruzzone, R.~Guan, and C.~Yang, ``Superpixel-level global and local similarity graph-based clustering for large hyperspectral images,'' \emph{IEEE Transactions on Geoscience and Remote Sensing}, vol.~60, pp. 1--16, 2022.

\bibitem{jampani18ssn}
V.~Jampani, D.~Sun, M.-Y. Liu, M.-H. Yang, and J.~Kautz, ``Superpixel samping networks,'' in \emph{European Conference on Computer Vision (ECCV)}, 2018.

\bibitem{achanta2012slic}
R.~Achanta, A.~Shaji, K.~Smith, A.~Lucchi, P.~Fua, and S.~S{\"u}sstrunk, ``Slic superpixels compared to state-of-the-art superpixel methods,'' \emph{IEEE transactions on pattern analysis and machine intelligence}, vol.~34, no.~11, pp. 2274--2282, 2012.

\bibitem{DEKM}
W.~Guo, K.~Lin, and W.~Ye, ``Deep embedded k-means clustering,'' in \emph{2021 International Conference on Data Mining Workshops (ICDMW)}, 2021, pp. 686--694.

\bibitem{SpectralNet}
U.~Shaham, K.~Stanton, H.~Li, B.~Nadler, R.~Basri, and Y.~Kluger, ``Spectralnet: Spectral clustering using deep neural networks,'' in \emph{Proc. ICLR 2018}, 2018.

\bibitem{guo2017improved}
X.~Guo, L.~Gao, X.~Liu, and J.~Yin, ``Improved deep embedded clustering with local structure preservation.'' in \emph{Ijcai}, vol.~17, 2017, pp. 1753--1759.

\bibitem{li2023model}
X.~Li, N.~Nadisic, S.~Huang, N.~Deligiannis, and A.~Pi{\v{z}}urica, ``Model-aware deep learning for the clustering of hyperspectral images with context preservation,'' in \emph{2023 31st European Signal Processing Conference (EUSIPCO)}.\hskip 1em plus 0.5em minus 0.4em\relax IEEE, 2023, pp. 885--889.

\bibitem{leiber2023benchmarking}
C.~Leiber, L.~Miklautz, C.~Plant, and C.~Böhm, ``Benchmarking deep clustering algorithms with clustpy,'' in \emph{2023 IEEE International Conference on Data Mining Workshops (ICDMW)}.\hskip 1em plus 0.5em minus 0.4em\relax IEEE, 2023, pp. 625--632.

\end{thebibliography}
\end{document}